\pdfoutput=1
\documentclass[10pt,twocolumn,letterpaper]{article}
\usepackage[pagenumbers]{cvpr}
\usepackage[T1]{fontenc}
\usepackage{microtype}
\usepackage{tikz}
\usetikzlibrary{arrows.meta,positioning}
\usepackage[ruled,linesnumbered]{algorithm2e}
\SetAlFnt{\footnotesize}
\SetAlCapSty{textnormal}
\SetAlCapFnt{\small}
\SetAlCapNameFnt{\small}
\SetAlgoCaptionSeparator{.}
\SetAlCapHSkip{0pt}
\SetArgSty{textnormal}

\SetCommentSty{smallitcomment}

\AtBeginDocument{%
  \crefname{algocf}{Algorithm}{Algorithms}%
  \Crefname{algocf}{Algorithm}{Algorithms}%
}

\definecolor{linkblue}{rgb}{0.21,0.49,0.74}
\usepackage{xr}
\usepackage[pagebackref,breaklinks,colorlinks,allcolors=linkblue]{hyperref}
\hypersetup{pdfauthor={Dawid Paj\k{a}k, Martin Bisson, Rodolfo Lima, Fran\c{c}ois Trudel},pdftitle={HiGS: A Hierarchical Architecture for Fast Differentiable 3D Gaussian Splatting}}

\title{HiGS: A Hierarchical Architecture for Fast Differentiable 3D Gaussian Splatting}

\author{
  Dawid Paj\k{a}k \quad Martin Bisson \quad Rodolfo Lima \quad Fran\c{c}ois Trudel \\
  NVIDIA
}

\begin{document}
\maketitle

\begin{abstract}
3D Gaussian Splatting (3DGS) makes one screen tile the cell of partitioning, the cell of rasterization, and the unit of parallel work.
Partitioning favors large tiles and rasterization small ones, and with one unit per tile the densest tiles set the frame time.
We present \emph{Hierarchically Tiled Gaussian Splatting} (HiGS), a unified architecture for rendering and reconstruction that gives each role its own granularity: Gaussians are partitioned over coarse \emph{macro-tiles}, rasterized over the fine render tiles within them, and processed in bounded depth-sorted batches.
A dense region issues more batches instead of one longer unit, so frame time follows the work in a view rather than its densest tile.
The same batches carry reconstruction: compositing supplies the transmittance and color needed to differentiate each one independently, leaving optimizer memory traffic as the dominant training cost.
We fuse the optimizer into the backward pass, halve the parameters and moments it stores and updates directly, and defer appearance updates outside the view.
HiGS renders scenes faster than every published method we evaluate, by the widest margin on views with uneven load, and reconstructs them faster and in less memory, at equal image quality and with exact front-to-back alpha compositing.

\end{abstract}

\section{Introduction}
\label{sec:intro}

3D Gaussian Splatting (3DGS)~\cite{kerbl2023gaussian} has become the leading approach to real-time photorealistic rendering of captured scenes, combining the fidelity of neural radiance fields~\cite{mildenhall2020nerf} with an explicit, differentiable representation that commodity GPUs rasterize at high throughput.
Reconstruction runs its rasterizer forward and backward tens of thousands of times, so rasterization cost governs both reconstruction and display.

\begin{figure}[t]
\centering
\includegraphics[width=\linewidth]{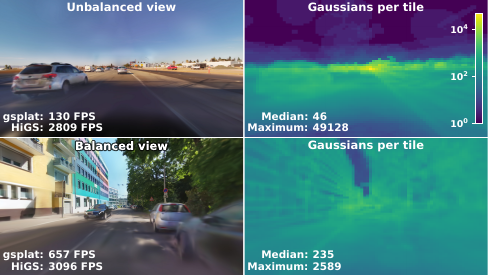}
\caption{\textbf{Per-tile load imbalance} on two driving views (1080p, $16\times16$ tiles), with the throughput of gsplat~\cite{ye2024gsplat} and of HiGS (\cref{sec:hierarchy}) on each. The median is over occupied tiles. The densest tiles gate the frame time of gsplat.}
\label{fig:imbalance}
\end{figure}

Most variants of 3DGS share one pipeline: Gaussians are projected to the screen, assigned to the tiles they overlap, sorted globally by tile and depth into one Gaussian list per tile, and rasterized with one thread block per tile.
The tile serves three conflicting roles: partition cell, rasterization cell, and unit of parallel work.
Partitioning favors large tiles, which reduce the number of (tile, Gaussian) pairs to bin and sort.
Rasterization favors small tiles, which reduce pixel--Gaussian evaluations outside the Gaussian's effective support.
At any size, a few dense tiles hold far more Gaussians than the rest and, with one thread block each, produce a \emph{tail effect} that dominates frame time (\cref{fig:imbalance}).
A single tile size cannot serve all three, and 3DGS pays with a global radix sort over tens of millions of pairs per frame at 4K and with a frame time set by the densest tile rather than by the work in the view.

Prior work lowers these costs without separating all three roles.
Pairs are reduced, models compressed, and single stages accelerated~\cite{hanson2024speedysplat,feng2024flashgs,wang2024adrgaussian,radl2024stopthepop,hahlbohm2026fastergs,fan2024lightgaussian,liao2025tcgs,schutz2025splatshop,mallick2024taming,durvasula2024distwar}, and methods that leave the pipeline pay with approximate compositing~\cite{hou2025sortfree,rueckert2025stochastic} or custom hardware~\cite{li2025gaurast,yang2025gsacc}.
Sorting at a coarser granularity than rasterization~\cite{lee2024gscore,jo2025gstg,ding2026sharpgs} shrinks the sort but leaves each tile one unbounded Gaussian list, as does rescheduling tiles~\cite{gui2024balanced}, and dividing the lists of the densest ones among thread blocks~\cite{zhang2026catgs} shortens the tail but keeps the global sort.
Each of these remedies changes one or two roles of the tile, and none carries the change into the backward pass of training.

We present HiGS, a unified architecture for commodity GPUs that gives each role its own granularity, with one unit of work for rendering and reconstruction.
Gaussians are binned and depth-sorted over coarse \emph{macro-tiles}, each a group of the fine \emph{render tiles} over which they are rasterized, and the unit of work is a \emph{(macro-tile, Gaussian-batch)} pair: a depth-sorted batch of bounded size, rasterized independently and then composited with the others in depth order.
Three consequences follow from this single decomposition.
Sorting splits into independent macro-tile segments.
Every batch, of a dense tile or not, fits in fast local storage, where it is loaded once and shared across its render tiles.
Dense macro-tiles issue proportionally more units, so no unit processes a dense tile's entire Gaussian list, which removes the cause of the tail effect.

The same units carry reconstruction, so both passes of a training step inherit this schedule.
Compositing leaves every batch with the transmittance in front of it and the color from it onward, and batches are differentiated as independently as they are rendered.
The optimizer then dominates training, and we reduce its memory traffic until it, too, scales with what a view sees: fused into the backward pass, it updates low-precision parameters and moments directly and defers the appearance updates of the Gaussians a view culls.
HiGS renders 1.5$\times$ to 5.6$\times$ faster than recent optimized rasterizers on standard benchmark scenes, and 6.8$\times$ to 20.0$\times$ faster on driving views with a strongly uneven load.
It reconstructs scenes 2.3$\times$ to 3.0$\times$ faster than the fastest published method we evaluate, with 1.9$\times$ less memory than the most memory-efficient one, at equal image quality.

Our contributions are the architecture and its three parts:
\begin{itemize}
    \item a \textbf{unified architecture} for rendering and reconstruction that separates partitioning, rasterization, and parallel work around one unit, a bounded depth batch of a macro-tile;
    \item a \textbf{renderer on these units}, with a segmented depth sort, load-balanced batch rasterization with inline visibility, and a stable half-precision (FP16) data path;
    \item a \textbf{backward pass over the same batches}, which differentiates each independently from boundary state left by compositing;
    \item a \textbf{bandwidth-lean optimizer} fused with backward projection, which updates stochastically rounded 16-bit state directly and defers appearance updates a view does not need.
\end{itemize}

\section{Related Work}
\label{sec:related}

\begin{figure*}[t]
\centering
\includegraphics[width=\textwidth]{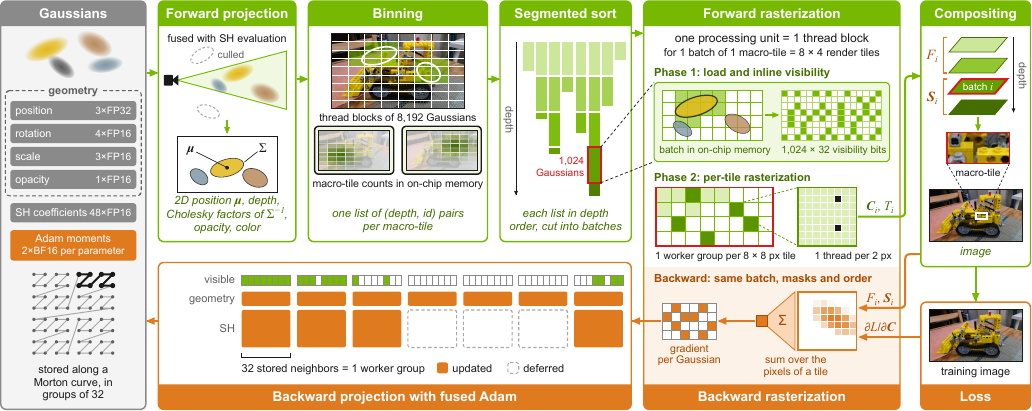}
\caption{\textbf{The HiGS pipeline}: the stages of every frame (green), those a training step adds (orange), and the Gaussians (grey). The rasterization panel shows one processing unit, shared by both passes: per pixel, it blends batch $i$ alone into color $\mathbf{C}_i$ and transmittance $T_i$. Compositing gives each batch the transmittance $F_i$ of the batches in front and the color $\mathbf{S}_i$ that it and those behind it contribute.}
\label{fig:pipeline}
\end{figure*}

Kerbl~\etal~\cite{kerbl2023gaussian} established the tile-based pipeline of 3D Gaussian Splatting, and gsplat~\cite{ye2024gsplat} popularized it in a widely adopted open-source library.
Most of the work that followed keeps the rendering model and lowers the cost of this pipeline.

\noindent\textbf{Pair reduction.}
Many pairs are false positives: a Gaussian's bounding box overlaps a tile the Gaussian does not visibly reach.
StopThePop~\cite{radl2024stopthepop} introduced opacity-aware bounds and an exact test of which tiles an ellipse reaches.
FlashGS~\cite{feng2024flashgs}, AdR-Gaussian~\cite{wang2024adrgaussian}, and Speedy-Splat~\cite{hanson2024speedysplat} derive similar bounds, the first and the last with exact tests of their own, and Faster-GS~\cite{hahlbohm2026fastergs} adopts them.
HiGS tests ellipse overlap during binning and rasterization, but most of its pair reduction comes from binning over macro-tiles.

\noindent\textbf{Model-side reduction.}
Fewer Gaussians also mean fewer pairs, and work complementary to ours shrinks the model: by pruning in Speedy-Splat and RadSplat~\cite{niemeyer2024radsplat}, by budgeted densification in Taming~3DGS~\cite{mallick2024taming}, by importance-weighted sampling in Mini-Splatting~\cite{fang2024minisplatting}, by multi-view consistent densification and pruning in FastGS~\cite{ren2026fastgs}, and by the quantization and codebook schemes of LightGaussian~\cite{fan2024lightgaussian}, EAGLES~\cite{girish2024eagles}, Compact-3DGS~\cite{lee2024compact3dgs}, and Compressed 3DGS~\cite{niedermayr2024compressed}.
Kerbl~\etal~\cite{kerbl2024hierarchical} build a level-of-detail hierarchy over the scene, rather than over the work of rendering it, for very large captures.

\noindent\textbf{Training acceleration.}
Several of these methods act during training, whose time is dominated by the backward pass and the optimizer.
The backward pass of 3DGS accumulates gradients per pixel and contends on atomic updates, which Taming~3DGS avoids by parallelizing over splats and DISTWAR~\cite{durvasula2024distwar} reduces by aggregating a warp's updates to the same Gaussian in registers before issuing one atomic per parameter.
LiteGS~\cite{liao2025litegs} reduces gradients within the one warp that owns a tile, and culls and updates Morton-ordered Gaussians in clusters.
On the optimizer side, Faster-GS pairs the per-splat backward pass with its own Adam implementation, which it can fuse into that pass, and reports up to $5\times$ faster training end to end.
GS-Scale~\cite{lee2026gsscale} keeps only geometry on the GPU and defers the Adam updates of the other parameters, held in host memory, while a Gaussian is outside the view frustum.
Other methods lower the work of a run rather than the cost of a step: DashGaussian~\cite{chen2025dashgaussian} schedules the rendering resolution and the Gaussian count, SAD-GS~\cite{lyu2026sadgs} shortens convergence by guiding densification with multiscale image structure, and Grendel~\cite{zhao2025grendel} distributes training over many GPUs.
These complement HiGS, which redesigns the training step itself.

\noindent\textbf{Rendering acceleration.}
In the forward pass, GPU implementations have tuned the pipeline stage by stage, most keeping its global sort and the screen tile as the unit of work.
FlashGS fuses the preprocessing kernels and pipelines the dependent Gaussian fetches in software, and TC-GS~\cite{liao2025tcgs} reformulates the alpha-blending exponent as a Tensor-Core matrix multiplication, reporting up to $2.2\times$ speedup when integrated into 3DGS.
Splatshop~\cite{schutz2025splatshop} narrows the sort keys by ordering visible splats by 32-bit depth before tile-fragment creation.
TileGS~\cite{tan2026tilegs} replaces the global sort by depth bins inside each tile and sorts only selected bins exactly.
Against uneven load, StopThePop balances the per-Gaussian stages, and AdR-Gaussian trains with a loss that evens the number of Gaussians per pixel.
In the rasterizer, Balanced 3DGS~\cite{gui2024balanced} counters load imbalance by letting thread blocks claim tiles at run time, shrinking the tile to four pixels, and spreading the Gaussians of a pixel across the threads of a warp.
CaT-GS~\cite{zhang2026catgs} divides the depth-ordered list of a tile above twice the mean load into slices that separate thread blocks rasterize and that are then composited in order, within the per-tile pipeline and its global sort.
For ultra-high resolutions, SHARP-GS~\cite{ding2026sharpgs} bins at a tile size that grows with resolution, culls against the finer tiles it rasterizes through a bitmask per batch of 32 Gaussians, and stores Gaussians in Morton order.
GS-TG~\cite{jo2025gstg}, an accelerator design that is also evaluated on a GPU, sorts over groups of tiles and rasterizes the original tiles through per-Gaussian bitmasks, and other hardware proposals~\cite{lee2024gscore,li2025gaurast,yang2025gsacc} leave the commodity GPU.
In HiGS one bounded batch is the unit of work for every list, in rendering and in both passes of reconstruction.

\noindent\textbf{Alternative formulations.}
The remaining work changes the rendering model itself, which HiGS keeps.
Exact front-to-back compositing is given up for a commutative weighted sum in Sort-free Gaussian Splatting~\cite{hou2025sortfree} and for stochastic transparency in StochasticSplats~\cite{rueckert2025stochastic}.
Others raise fidelity through per-ray ordering in StopThePop, band-limited primitives~\cite{yu2024mipsplatting}, and perspective-correct evaluation with hybrid transparency~\cite{hahlbohm2025hybrid}, or extend the model to secondary rays~\cite{moenneloccoz2024gaussianrt} and distorted cameras~\cite{wu20253dgut}.

\section{Rendering Architecture}
\label{sec:arch}

The HiGS pipeline (\cref{fig:pipeline}, see \cref*{S-app:implementation} for implementation details) is a hierarchy of work units, fed by fused projection and spherical-harmonics (SH) evaluation as in recent rasterizers~\cite{feng2024flashgs,hahlbohm2026fastergs}, which emits each Gaussian's screen-space center, depth, inverse-covariance factors, color, and visibility.

\subsection{Macro-tile hierarchy and work decomposition}
\label{sec:hierarchy}

The hierarchy is built over the standard rasterization primitive, the \emph{render tile}, here $8\times8$ pixels.
HiGS groups $M{=}32$ adjacent render tiles into a \emph{macro-tile}, arranged as $8\times4$ tiles spanning $64\times32$ pixels.
Binning and sorting produce one depth-ordered Gaussian list per macro-tile (\cref{sec:binning}), while visibility and blending operate per render tile (\cref{sec:fused-raster}).
Within each macro-tile, Gaussian data is loaded into local storage once and reused across all $M$ member render tiles, amortizing the fetch over fine-grained output.

Each list is split into \emph{Gaussian batches} of at most $B_r{=}1{,}024$ Gaussians, each (macro-tile, batch) pair forming an independent \emph{processing unit}.
The bound holds for every list, dense or not, because it is what lets a batch reside in local storage.
A macro-tile with $n$ Gaussians thus spawns $\lceil n/B_r \rceil$ units.
Within each processing unit, render tiles are claimed dynamically by \emph{worker groups} from a shared \emph{tile work queue} $\mathcal{Q}$, and a per-worker \emph{scheduling buffer} $\mathcal{R}$ absorbs the per-tile variability in visible-Gaussian count so the blending pipeline runs at a fixed rate.
The schedule is therefore density-proportional rather than screen-area-proportional: compute scales with Gaussian count, and no single render tile can gate the frame the way the slowest tile gates a one-thread-group-per-tile assignment (\cref{fig:imbalance-time}).

Every level of this hierarchy maps onto a level of the GPU (\cref{fig:pipeline}).
A processing unit is one thread block whose batch resides in on-chip memory, our local storage; a worker group is $P{=}32$ threads that execute in lockstep; and each thread keeps its pixel accumulators in registers.
Preallocated macro-tile lists and batch partial buffers let units run independently, with storage growing with the batch count.

\subsection{Partitioning and segmented depth sort}
\label{sec:binning}
\label{sec:sort}

Binning constructs a depth-keyed Gaussian list per macro-tile.
List lengths are data-dependent, so building the lists in parallel requires reconciling write positions without serializing through global atomics.
Kerbl~\etal~\cite{kerbl2023gaussian} leave this to the sort, whose composite (tile, depth) key groups unordered records by tile and orders them by depth, which makes it global.

Our binning produces the per-macro-tile lists \emph{during} binning, leaving only in-segment ordering for the sort.
Gaussians are processed in partitioning batches of $B_p{=}8{,}192$, each handled cooperatively by one group of threads that keeps a private histogram in local storage, one counter per macro-tile of the frame, and accumulates overlap counts with local rather than global atomics.
The histogram fits because the frame's macro-tile count is modest, roughly 1{,}000 entries at 1080p and 4{,}000 at 4K; a render-tile histogram would carry 30{,}000 to 130{,}000 entries, exceed local storage by an order of magnitude, and force a global-atomic path.
A hierarchical prefix scan turns the batch-local counts into write offsets, and a fill pass repeats the batches and writes (depth, Gaussian-id) pairs directly into each macro-tile's range.

Gaussians are paired with macro-tiles by a conservative per-row scan of the opacity-threshold ellipse, following AccuTile~\cite{hanson2024speedysplat} one level coarser.
HiGS thus rasterizes over $8\times8$ tiles yet averages only 1.51M pairs at 1080p and 2.21M at 4K on Mip-NeRF~360, $2.7\times$ and $4.3\times$ fewer than gsplat's opacity-aware bounding boxes at $16\times16$ tiles, which already form $2.5\times$ and $3.1\times$ fewer than at $8\times8$.
Macro-tile binning gives over half of our reduction, the ellipse test the rest.

Each macro-tile is then one segment of a \emph{segmented sort}, and segments are independent by construction.
A segment already identifies its macro-tile, so its key is the 32-bit depth alone, rather than gsplat's 64-bit composite key.
Removing tile bits shortens the radix sort and shrinks its records from 12 to 8~bytes.
Splatshop~\cite{schutz2025splatshop} narrows the key within the global sort, whereas here the tile dimension disappears from it.
As resolution grows, the render-tile pair count grows but segments stay bounded, because more macro-tiles distribute the Gaussians into more, smaller segments.
A general-purpose segmented sort still slows down as the segments multiply.
We therefore sort with a pipeline of six stages that run without host synchronization: a cascade of four block radix sorts~\cite{merrill2010revisiting}, one per class of segment size, and, for the segments above the largest class, a sort of fixed-size runs followed by their merge with merge path~\cite{green2012gpu}.
The sort then stays in the same time range from 1080p to 4K (\cref{sec:results-perf}).

\subsection{Batch rasterization and compositing}
\label{sec:fused-raster}

3DGS materializes explicit per-tile Gaussian lists before rasterization can begin.
Our rasterizer replaces that stage with visibility determination inline with the Gaussian load, at low marginal cost.
Each processing unit consists of $W$ worker groups of $P$ threads and handles its (macro-tile, batch) pair in two phases, followed by a compositing pass shared by all units (\cref{fig:pipeline}).

\noindent\textbf{Phase~1: load and inline visibility.}
The unit cooperatively loads its batch into local storage in mini-batches of $P$ Gaussians, one per thread.
During the load, each Gaussian's opacity-threshold ellipse is tested for overlap with all $M$ render tiles, producing a per-Gaussian tile mask.
GS-TG~\cite{jo2025gstg} materializes these masks before rasterization, which on a GPU serializes their generation with the sort, and SHARP-GS~\cite{ding2026sharpgs} likewise builds one per batch of 32 Gaussians; computing them during the cooperative load avoids that memory round trip.
The load identifies which tiles each Gaussian overlaps, but blending needs the reverse, which Gaussians each tile sees.
A transpose of each mini-batch's $P{\times}M$ bit matrix converts one into the other: every render tile then holds a $P$-bit mask per mini-batch, and their concatenation is its visibility over the batch.

\noindent\textbf{Phase~2: per-tile rasterization.}
Worker groups claim active render tiles from $\mathcal{Q}$, so no group idles while tiles remain.
For a claimed tile, the group walks its masks and extracts only the visible Gaussians from local storage.
The scheduling buffer $\mathcal{R}$ accumulates their indices and drains them in batches of $B_d{=}64$ through an unrolled blending core, which decouples variable-rate mask iteration from fixed-rate blending arithmetic.
After each drain, a unanimous vote tests whether every pixel of the tile has reached transmittance saturation; if so, the group claims the next tile.

\noindent\textbf{Compositing.}
A macro-tile's list may span several batches processed by independent units, so per-tile results are partial: batch $i$ produces, per pixel, a local color $\mathbf{C}_i$ and a local transmittance $T_i$, blended over its own Gaussians only.
Front-to-back alpha blending is associative over contiguous depth ranges~\cite{porter1984compositing}, as Balanced 3DGS and CaT-GS also use, so with $F_i=\prod_{h<i}T_h$ the transmittance in front of batch $i$, the pixel color is
\begin{equation}
    \mathbf{C} = \sum_i F_i\,\mathbf{C}_i ,
    \label{eq:compose}
\end{equation}
plus the background weighted by the final transmittance.
Each unit writes its partials only for tiles that received Gaussians, flagged in a per-batch active-tile mask, and a compositing pass walks each render tile's batches in depth order, accumulating \cref{eq:compose}.
The pass is sequential across the batches of a tile but not a bottleneck: it reads a compact stream of partials, skips the remaining batches once the tile saturates, and takes 0.01 to 0.03~ms at 1080p.
Every pixel is thus composited front to back in the depth order of 3DGS and by its alpha-blending recurrence, which is the sense in which we call HiGS's compositing \emph{exact}.

\subsection{Numerical representation and storage order}
\label{sec:numerical}

The rasterizer stores and evaluates projected attributes in FP16.
The motivation is memory rather than arithmetic: packed FP16 and single-precision (FP32) share the same math throughput on the GPUs we target, but 16-bit attributes halve the footprint in memory and in local storage, so a processing unit loads twice as fast, keeps twice as many Gaussians resident, and carries half the accumulator state per thread.
Two hazards arise in the per-pixel Mahalanobis distance $q = (\mathbf{p} - \boldsymbol{\mu})^\top \Sigma^{-1} (\mathbf{p} - \boldsymbol{\mu})$, for pixel $\mathbf{p}$, projected center $\boldsymbol{\mu}$, and screen-space covariance $\Sigma$, and HiGS addresses both in the data representation rather than per pixel.

Cancellation arises when $\Sigma^{-1}$ is stored as conic coefficients, whose quadratic form has a mixed-sign cross term.
HiGS stores the Cholesky factor of $\Sigma^{-1} = L L^\top$ instead, a lower-triangular $L$ with entries $l_0$, $l_1$, and $l_2$, so that with the offset $(dx, dy) = \boldsymbol{\mu} - \mathbf{p}$
\begin{equation}
    q = (l_0\,dx + l_1\,dy)^2 + (l_2\,dy)^2 ,
    \label{eq:cholesky}
\end{equation}
a sum of squares that is non-negative regardless of rounding.
Overflow arises when coordinates carry their absolute screen-space magnitudes into these products; a unit narrows positions to FP16 only relative to its macro-tile center, in render-tile units.
TC-GS~\cite{liao2025tcgs} identifies the same hazard and shifts coordinates per tile; here one shift at load serves all render tiles of the macro-tile.
For rendering, SH coefficients can additionally be packed into 32~bytes per Gaussian, a sixth of their FP32 size, through YCoCg decorrelation~\cite{malvar2008lifting} and non-uniform quantization decoded inline~\cite{nvidia2025ptx}.

Storage order also affects memory traffic.
Projection reads SH coefficients, and binning reads projected attributes, only for the Gaussians that a view does not cull.
Morton ordering of Gaussian means, also used by SHARP-GS and LiteGS, groups culled Gaussians into contiguous runs.
Their data is skipped in whole memory sectors rather than fetched alongside visible neighbors, shortening the frame (\cref{sec:results-perf}).
The order is independent of the per-frame depth sort, and the same coherence carries the optimizer (\cref{sec:deferred}).

\section{Reconstruction}
\label{sec:training}

A training step renders one view with the pipeline of \cref{sec:arch}, differentiates the image loss through it, and updates the parameters.
We retain the objective and densification policy of 3DGS as implemented in gsplat.
The independent processing units carry the backward pass, after which optimizer memory traffic dominates the step, and we reduce it until appearance updates scale with what a view sees, not with the scene size.

\subsection{Differentiating independent batches}
\label{sec:backward}

The gradient of a Gaussian's opacity $\alpha$ at a pixel depends on the transmittance in front of it and the color composited behind it.
3DGS recovers them by walking each tile's list back to front and dividing transmittance by $1-\alpha$, StopThePop~\cite{radl2024stopthepop} walks front to back and subtracts from the final color, and Taming~3DGS~\cite{mallick2024taming} and Faster-GS walk front to back from per-pixel state stored every 32 Gaussians.
None fits HiGS directly, because a tile's list spans independently rendered batches, and state stored per tile at that density would forfeit the memory economy of shared loading.

We store this state once per batch.
The training forward accumulates color in FP32, keeping blend weights and transmittance in FP16, and the compositing pass of \cref{sec:fused-raster} completes the \emph{boundary state} of every batch.
Its front-to-back walk already forms the transmittance $F_i$ in front of batch $i$; a second walk, back to front, replaces each color partial by the suffix
\begin{equation}
    \mathbf{S}_i=\sum_{h\geq i}F_h\,\mathbf{C}_h ,
    \label{eq:suffix}
\end{equation}
the color composited from batch $i$ onward.
Both overwrite the partials in place, so the boundary state costs no storage beyond the 14-byte partial buffer per pixel and batch.
A backward processing unit then needs no other batch: it differentiates its (macro-tile, batch) pair from $F_i$, $\mathbf{S}_i$, and the batch.

Inside the unit, the forward pass is evaluated again rather than inverted.
The unit loads the same batch into local storage, reuses the visibility masks saved by the forward, and each worker group blends its tile front to back with the forward's arithmetic, drain boundaries, and saturation test, starting from a local transmittance of one.
This reproduces the rounded blend weight $\widehat{w}_j=\alpha_j t_j$ of every Gaussian $j$ exactly, where $\alpha_j$ is its opacity at the pixel and $t_j$ the local transmittance in front of it, which a division of FP16 transmittances cannot do near saturation.
With $\mathbf{c}_j$ the color of Gaussian $j$, $\mathbf{g}=\partial\mathcal{L}/\partial\mathbf{C}$ the gradient of the loss $\mathcal{L}$ at the pixel, $\mathbf{g}_i=F_i\,\mathbf{g}$, background $\mathbf{b}$, and final transmittance $T_{\mathrm{end}}$, the group initializes a scalar residual $r=\mathbf{g}^\top(\mathbf{S}_i+T_{\mathrm{end}}\mathbf{b})$, the loss-weighted color at and behind the batch, and for each Gaussian in turn computes
\begin{align}
    r &\leftarrow r-\widehat{w}_j\,\mathbf{g}_i^\top\mathbf{c}_j, \label{eq:residual}\\
    \frac{\partial\mathcal{L}}{\partial\alpha_j} &= t_j\,\mathbf{g}_i^\top\mathbf{c}_j-\frac{r}{1-\alpha_j},
    \qquad
    \frac{\partial\mathcal{L}}{\partial\mathbf{c}_j}=\widehat{w}_j\,\mathbf{g}_i ,
    \label{eq:backward}
\end{align}
so that $r$ always holds the contribution behind the current Gaussian.

The remaining chain rule, from $\alpha_j$ to the projected center, the Cholesky factors, and the opacity, is evaluated per pixel and reduced within the worker group, so that five two-word atomic additions per Gaussian and tile update the nine accumulators of its geometry and color gradients and no per-pixel gradient buffer exists.

\subsection{Optimizer fused with projection}
\label{sec:optimizer}

Accelerating rasterization exposes Adam's bandwidth cost: every step reads and writes all parameters and their two moments, a state of 708~bytes per Gaussian in FP32 with degree-3 SH.
Behind our rasterizer, PyTorch's Adam takes longer than the forward and backward passes together (\cref{sec:ablations}), and the three measures discussed below reduce its traffic.

The first measure removes the gradients from memory.
As in Faster-GS, the update is applied inside the backward pass: backward projection assigns one thread per Gaussian, which differentiates projection and SH evaluation and therefore holds the gradient of the Gaussian's geometry in registers, together with both factors of every SH gradient.
It applies Adam there, the threads of a worker group updating their SH coefficients cooperatively, and writes each parameter and moment once.
The only gradients that ever reach memory are the rasterizer's accumulators, 40~bytes per Gaussian, which are needed because many tiles contribute to one Gaussian.

\subsection{Low-precision parameters and moments}
\label{sec:precision}

Fusion removes the gradient traffic but not the state.
The second measure halves it: positions stay in FP32, all other parameters (rotation, log-scale, opacity logit, and SH coefficients) are stored in FP16, and both Adam moments in bfloat16 (BF16)~\cite{kalamkar2019bfloat16}, for 360~bytes per Gaussian instead of 708.
These arrays are the optimizer's own state, not a rounded copy of FP32 parameters, so the renderer reads exactly what the optimizer wrote.
Projection, its derivative, and Adam arithmetic remain in FP32; only the stores are narrowed.

Narrowing a store by rounding to nearest discards every update below half a unit in the last place, which the small learning rates of splatting optimization make common.
We round stochastically~\cite{gupta2015limited}, as Zamirai~\etal~\cite{zamirai2021bfloat16} do for the weights of BF16 training, and round the moments as well: the stored value is one of the two representable neighbors of the exact result, chosen with probability proportional to proximity, so the stored value equals the exact one in expectation.
With rounding to nearest, reconstruction quality degrades measurably; with stochastic rounding it matches FP32 storage (\cref{sec:ablations}).

\subsection{Visibility-proportional appearance updates}
\label{sec:deferred}

What remains is mostly appearance: SH coefficients are 48 of a Gaussian's 59 scalars, yet a training view of the \emph{garden} scene of Mip-NeRF~360 culls 66\% of the Gaussians on average, and their gradients are zero.
Skipping them, as Taming~3DGS suggests, Faster-GS evaluates, and LiteGS does for whole clusters, is not equivalent to updating them, since Adam's first moment keeps moving a parameter after its gradient vanishes, which Ding~\etal~\cite{ding2026decouple} analyze as the implicit update.
A zero-gradient step depends only on the stored moments and the step index, however.
As sparse online learning does for regularization~\cite{langford2009truncated,duchi2009fobos}, we apply such steps lazily: a culled coefficient stays untouched until a view needs it and then takes all the steps it missed in closed form.

Let $(m,v)$ be the moments of such a coefficient $\theta$ after its last update at step $a$.
Over $k$ zero-gradient steps Adam, with decay rates $\beta_1$ and $\beta_2$, decays them to $(\beta_1^k m,\ \beta_2^k v)$, and with learning rate $\eta$ and $\rho=\beta_1/\sqrt{\beta_2}$ we move $\theta$ by
\begin{align}
    \Delta\theta &= -\,\frac{\eta\,m\,\Phi}{\sqrt{v}+\varepsilon_A\,\Phi/\Psi}, \label{eq:catchup}\\
    \Phi &= \sum_{j=1}^{k}\rho^{\,j}\frac{\sqrt{1-\beta_2^{\,a+j}}}{1-\beta_1^{\,a+j}},
    \qquad
    \Psi = \sum_{j=1}^{k}\frac{\beta_1^{\,j}}{1-\beta_1^{\,a+j}}, \nonumber
\end{align}
where $\Phi$ and $\Psi$ are the sums of the bias-corrected step factors in the regimes where Adam's $\varepsilon_A$ is negligible against $\sqrt{v}$ and where it dominates.
For the constant learning rate of SH coefficients, \cref{eq:catchup} equals Adam's displacement in both regimes and for $k{=}1$, and between them exceeds it by less than 0.05\% from step 100 on.
Each sum comes from two entries of a table built once per run, so the lag needs no bound: only the step of the last update is kept, and a coefficient that no view needs is neither read, written, nor counted.
The forward pass applies \cref{eq:catchup} in registers when it reads lagging coefficients, and the fused optimizer applies it with the current step and stores the result once.
GS-Scale~\cite{lee2026gsscale} also defers the Adam updates of its host-side parameters, with a restoration exact for $\varepsilon_A{=}0$ and a four-bit counter that forces an update at least every 16~steps, whereas lags on \emph{garden} have a median of 2~steps and a 99th percentile of 38.

We record the last update once per group of $P{=}32$ consecutive Gaussians, which one worker group of our optimizer updates.
A group with no visible Gaussian skips its SH update, a worker group uniform decision that leaves its 288~bytes per Gaussian of coefficients and moments untouched, and any other group updates all of its members.
Geometry is updated at every step, since it decides visibility.
A record per Gaussian would skip little more, because storage is in the Morton order that already shortens rendering (\cref{sec:numerical}): on \emph{garden}, 98\% of culled Gaussians lie in fully culled groups, even after densification ends.
In the order that gsplat's densification leaves, with new Gaussians appended, 11\% do.
Faster-GS reorders Gaussians every 5{,}000 iterations to reduce cache misses and divergent control flow, and LiteGS after densification steps.
Here the order decides how much appearance work is deferred, and every densification step restores it in one gather of all per-Gaussian arrays, with each copy or pair of split children written in its parent's place.
The optimizer then moves 375~bytes per Gaussian and step, against 1{,}416 for a dense FP32 update, which reads and writes the whole state once (see \cref*{S-app:reconstruction} for detailed derivations).

\section{Experiments}
\label{sec:results}

\subsection{Setup}
\label{sec:setup}

We evaluate our method using a PC with NVIDIA RTX PRO 6000 Blackwell GPU and AMD Ryzen 7 7800X3D CPU.
We compare against methods with a public implementation.

\noindent\textbf{Reconstruction.}
One training harness gives every method the same initialization procedure, training-view sequence, loss, schedules, and 3DGS densification policy.
The renderer, backward pass, optimizer, and storage order differ.
We compare against 3DGS, gsplat, and Faster-GS with its fused optimizer.
All methods use an opacity learning rate of 0.025.
Each trains for 30{,}000 steps on the 13 scenes commonly used to evaluate 3DGS, at the standard resolution of each dataset: the nine of Mip-NeRF~360~\cite{barron2022mipnerf360}, two of Tanks and Temples~\cite{knapitsch2017tanks}, and two of Deep Blending~\cite{hedman2018deepblending}.
We report the time of the training loop, without the startup that decodes the training images and caches them in GPU memory, peak GPU memory without that cache, and PSNR (in dB), SSIM~\cite{wang2004ssim}, and LPIPS~\cite{zhang2018lpips} with AlexNet features on the test set.

\noindent\textbf{Rendering.}
We render the publicly released 3DGS pre-trained models of seven Mip-NeRF~360 scenes (SH degree~3, 1.2M to 6.1M Gaussians) from every test camera at 1080p and 4K, and report each scene's median over all cameras' timed frames, using 100 frames per camera after 10 warm-up frames.
CUDA events time all GPU kernels that render a frame.
For uneven load we add 120 driving views at 1080p, one per second of six clips of the PhysicalAI Autonomous Vehicles dataset reconstructed by NuRec~\cite{nvidia2025nurec}, each a model of 2.5M Gaussians.
HiGS runs in its default configuration: $8\times8$ render tiles, Morton-ordered storage, and 32-byte SH.
Each prior rasterizer runs the default configuration of its public release at $16\times16$ tiles ($16\times12$ for Faster-GS), with the same tile-level depth ordering as ours and no antialiasing.

\subsection{Reconstruction}
\label{sec:reconstruction-results}

\begin{table}[t]
\centering
\footnotesize
\setlength{\tabcolsep}{3pt}
\begin{tabular}{@{}llrrrrr@{}}
\toprule
Dataset & Method & Time (s) & Mem.\ (GiB) & PSNR & SSIM & LPIPS \\
\midrule
Mip-NeRF & 3DGS      & 650.9 & 3.76 & 27.81 & .8245 & .1693 \\
360      & gsplat    & 309.4 & 3.87 & 27.82 & .8254 & .1671 \\
         & Faster-GS & 130.5 & 5.73 & 27.81 & .8253 & .1676 \\
         & HiGS      & 49.3 & 2.02 & 27.81 & .8254 & .1666 \\
\midrule
Tanks and & 3DGS      & 399.8 & 2.23 & 23.39 & .8386 & .1409 \\
Temples   & gsplat    & 213.8 & 2.30 & 23.41 & .8389 & .1402 \\
          & Faster-GS & 85.4 & 3.41 & 23.41 & .8389 & .1405 \\
          & HiGS      & 36.5 & 1.19 & 23.40 & .8387 & .1398 \\
\midrule
Deep     & 3DGS      & 765.2 & 3.73 & 29.83 & .9076 & .1650 \\
Blending & gsplat    & 293.8 & 3.86 & 29.82 & .9075 & .1644 \\
         & Faster-GS & 118.6 & 5.74 & 29.84 & .9076 & .1645 \\
         & HiGS      & 40.1 & 2.01 & 29.80 & .9072 & .1632 \\
\bottomrule
\end{tabular}
\caption{\textbf{Reconstruction}, mean over the scenes of each dataset and over ten runs (three for time and memory).}
\label{tab:training}
\end{table}

Across the three datasets HiGS completes a reconstruction 2.3$\times$ to 3.0$\times$ faster than the next fastest method and 5.9$\times$ to 7.3$\times$ faster than gsplat (\cref{tab:training}).
It also uses 1.9$\times$ less memory than the most memory-efficient alternative, a ratio close to that of the optimizer state, which the optimizer updates directly, with no FP32 copy (\cref{sec:precision}).
Dataset-average quality is equivalent: the PSNR spread across methods is smaller than HiGS's variation across repeated runs.
HiGS is the only method of \cref{tab:training} whose rasterizer issues work in proportion to Gaussian density in both passes of a training step, so reconstruction inherits the renderer's gains.
The rest of its advantage follows from \cref{sec:training}: the other methods update the appearance state of every Gaussian at every step, and those without a fused optimizer also move a full parameter gradient.
Faster-GS, the one baseline with a fused optimizer, is accordingly the next fastest, and \cref{sec:ablations} measures both parts on \emph{garden} (see \cref*{S-app:results} for per stage breakdown).

\subsection{Rendering}
\label{sec:results-perf}

\begin{table}[t]
\centering
\footnotesize
\setlength{\tabcolsep}{5pt}
\begin{tabular}{@{}lrrrr@{}}
\toprule
& \multicolumn{2}{c}{Throughput (FPS)} & \multicolumn{2}{c}{HiGS speedup} \\
\cmidrule(lr){2-3}\cmidrule(lr){4-5}
Method & 1080p & 4K & 1080p & 4K \\
\midrule
HiGS & 2756 & 1533 & --- & --- \\
SHARP-GS & 1250 & 1055 & 2.2$\times$ & 1.5$\times$ \\
Faster-GS & 949 & 578 & 2.9$\times$ & 2.7$\times$ \\
FlashGS & 893 & 670 & 3.1$\times$ & 2.3$\times$ \\
TC-GS & 765 & 499 & 3.6$\times$ & 3.1$\times$ \\
Speedy-Splat & 643 & 385 & 4.3$\times$ & 4.0$\times$ \\
StopThePop & 573 & 293 & 4.8$\times$ & 5.2$\times$ \\
gsplat & 541 & 275 & 5.1$\times$ & 5.6$\times$ \\
3DGS & 286 & 102 & 9.6$\times$ & 15.0$\times$ \\
\bottomrule
\end{tabular}
\caption{\textbf{Rendering} throughput, $1000$ over the mean of the per-scene median frame times in milliseconds.}
\label{tab:fps-comparison}
\end{table}

HiGS is the fastest rasterizer on every scene at both resolutions (\cref{tab:fps-comparison}).
Its lead narrows with how much of the per-tile pipeline a method changes: 4.0$\times$ to 5.6$\times$ over gsplat, Speedy-Splat, and StopThePop, which tighten the tile bounds of 3DGS, 2.3$\times$ to 3.6$\times$ over FlashGS, Faster-GS, and TC-GS, which rework its stages, and least over SHARP-GS, the one baseline that bins coarser than it rasterizes.
That proximity does not hold as the load grows: on \emph{garden} at a budget of 90M Gaussians, SHARP-GS falls behind FlashGS, and HiGS leads it by 3.4$\times$ at 1080p and 3.3$\times$ at 4K.

On Mip-NeRF~360 a HiGS frame averages 0.36~ms at 1080p, of which rasterization and compositing take 38\% to 52\%, binning with sort 25\% to 43\%, and projection with SH evaluation 9\% to 32\%, growing with the Gaussian count.
The depth sort itself takes 0.03 to 0.06~ms at either resolution, against 0.44 to 1.10~ms at 4K for gsplat's global radix sort, and it is 4.6$\times$ to 6.0$\times$ faster at 4K than a general-purpose segmented sort over the same segments.
Morton ordering, a one-time permutation of a model, lowers the frame time by 24\% at 1080p and 18\% at 4K, and in the stored order of the pretrained models HiGS is still 3.9$\times$ and 4.6$\times$ faster than gsplat.
SH packing lowers it by 11\% and 7\%, and without packing HiGS still leads every rasterizer of \cref{tab:fps-comparison}, SHARP-GS by 2.0$\times$ and 1.4$\times$.

\begin{figure}[t]
\centering
\includegraphics[width=\linewidth]{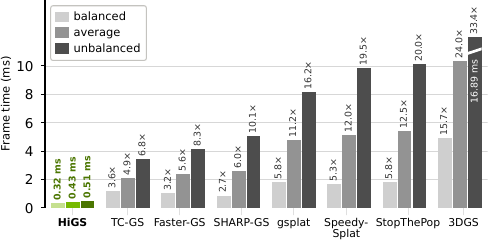}
\caption{\textbf{Frame time under load imbalance} on the 120 driving views: the mean over the 30 most balanced views, all views, and the 30 most unbalanced, ranked by the ratio of the densest tile to the median occupied one (\cref{fig:imbalance}). Labels give HiGS's frame time and its speedup over each rasterizer; the bar cut short carries its time. FlashGS, which does not complete the benchmark, is omitted.}
\label{fig:imbalance-time}
\end{figure}

\cref{fig:imbalance-time} shows what one unit per tile costs on uneven loads.
From the most balanced driving views to the most unbalanced, the tile--Gaussian pairs grow by 1.5$\times$, the frame time of HiGS by 1.6$\times$, and that of the seven baselines by 3.0$\times$ to 5.9$\times$.
Over all views, the frame time of HiGS follows the work in the view: a linear fit to the pair count explains 90\% of its variance, and one to the Gaussian count of the densest tile 53\%.
The frame time of the baselines follows the single densest tile, whose Gaussian count explains 87\% to 98\% of their variance and the pair count 30\% to 40\%.
Coarser binning does not change this: SHARP-GS, the closest in \cref{tab:fps-comparison}, is the one most tied to the densest tile, as each of its tiles remains one unit.
Reconstruction inherits this balance by construction, as the same units carry both passes of a training step (\cref{sec:backward}).

With the models fixed, HiGS agrees with gsplat to 67.2~dB PSNR, and to 55.6~dB with packed SH.
Compositing is exact, so the differences come from FP16 arithmetic, a saturation test every $B_d$ Gaussians, and an opacity limit of one instead of 0.99.
Against the test set, HiGS and six baselines lie within 0.03~dB in PSNR and 0.001 in SSIM and LPIPS of one another (see \cref*{S-app:results} for additional details).

\subsection{Ablations}
\label{sec:ablations}

\begin{table}[t]
\centering
\footnotesize
\setlength{\tabcolsep}{3.5pt}
\begin{tabular}{@{}lrrr@{}}
\toprule
Configuration & Time (s) & Mem.\ (GiB) & PSNR (dB) \\
\midrule
separate optimizer, FP32             & 194.4 & 4.14\,{\scriptsize$\pm$\,.02} & 27.83\,{\scriptsize$\pm$\,.03} \\
+ spatially ordered storage          & 176.8 & 3.99\,{\scriptsize$\pm$\,.01} & 27.82\,{\scriptsize$\pm$\,.04} \\
+ BF16 moments, to nearest           & 136.7 & 3.20\,{\scriptsize$\pm$\,.01} & 27.68\,{\scriptsize$\pm$\,.02} \\
+ stochastic moment rounding         & 137.0 & 3.18\,{\scriptsize$\pm$\,.01} & 27.82\,{\scriptsize$\pm$\,.05} \\
+ fused optimizer                    & 100.6 & 3.38\,{\scriptsize$\pm$\,.01} & 27.83\,{\scriptsize$\pm$\,.03} \\
+ FP16 parameters, to nearest        & 82.0 & 2.69\,{\scriptsize$\pm$\,.01} & 27.77\,{\scriptsize$\pm$\,.06} \\
+ stochastic parameter rounding      & 82.9 & 2.72\,{\scriptsize$\pm$\,.01} & 27.81\,{\scriptsize$\pm$\,.04} \\
+ deferred appearance (HiGS)         & 61.5 & 2.72\,{\scriptsize$\pm$\,.01} & 27.82\,{\scriptsize$\pm$\,.05} \\
\midrule
HiGS, no stochastic rounding         & 60.1 & 2.65\,{\scriptsize$\pm$\,.01} & 27.45\,{\scriptsize$\pm$\,.03} \\
HiGS, storage not ordered            & 91.2 & 2.72\,{\scriptsize$\pm$\,.01} & 27.84\,{\scriptsize$\pm$\,.05} \\
\bottomrule
\end{tabular}
\caption{\textbf{Ablation} of the reconstruction pipeline on the \emph{garden} scene, mean and standard deviation over ten runs. Each row adds one element of \cref{sec:training} to the row above, starting from the HiGS rasterizer with a separate FP32 Adam and gsplat's densification. The last two remove one element from HiGS.}
\label{tab:ablation}
\end{table}

The first row of \cref{tab:ablation} differs from gsplat in the rasterizer alone: it trains \emph{garden} in 194.4~s, against 404.8~s for gsplat, and the optimizer takes 59\% of its loop.
Five elements then shorten training to 61.5~s, and two restore quality.
Relative to the row above, our densification, which keeps storage spatially ordered (\cref{sec:deferred}), removes 9\% of training time, our separate Adam with BF16 moments 23\%, fusing the optimizer 27\% at 0.20~GiB more peak memory, FP16 parameters 18\%, and deferring the appearance updates of culled groups the last 26\%.
In all, training is 3.2$\times$ faster in 34\% less memory, at a PSNR 0.01~dB from that of the first row.
Rounded to nearest, the moments lose 0.14~dB and the parameters 0.06~dB, which stochastic rounding recovers to within the variation between runs, and HiGS without it loses 0.37~dB, more than the two losses combined.
Without spatially ordered storage HiGS takes 48\% longer, where Faster-GS reports 2\% to 6\% for training without its periodic reordering.
The order does more here than improve memory locality: 90\% of the added time is in the backward pass, where the fused optimizer loses the gain of deferral once few groups are fully culled (\cref{sec:deferred}), and the rest is in the forward pass (see \cref*{S-app:results} for additional results).

\section{Conclusion}
\label{sec:conclusion}

HiGS starts from one observation: partitioning, rasterization, and parallel work need not share a tile.
With a granularity for each, sorting shrinks to small independent segments, data is loaded once and shared, and no single dense tile gates the frame, all as consequences of one decomposition and not as separate optimizations.
Reconstruction inherits the same units and the rule behind them, to spend work only where a view needs it, down to the optimizer's memory traffic.
The result renders and reconstructs faster than every published method we evaluate, and reconstructs in less memory, with exact compositing and equal image quality.

This parallelism is bought with memory and bandwidth.
Every batch reserves and writes its own output and is rasterized even where opaque foreground hides it, so in a degenerate scene whose Gaussians all cover the whole frame, these costs grow with every hidden layer.
Captured scenes are far from that extreme, and in all our tests the balanced schedule outweighs the work that compositing discards, most of all on views with a strongly uneven load (\cref{fig:imbalance-time}).
Because a unit needs only its batch and the boundary state of its tiles, and appearance parameters rest untouched while out of view, we see scenes beyond one GPU's memory, spread over several GPUs or held partly out of core, as the natural next step.

{\small
\bibliographystyle{ieeenat_fullname}
\bibliography{main}
}

\clearpage
\maketitlesupplementary
\appendix
\setcounter{figure}{0}
\setcounter{table}{0}
\setcounter{algocf}{0}
\renewcommand{\thefigure}{S\arabic{figure}}
\renewcommand{\thetable}{S\arabic{table}}
\renewcommand{\thealgocf}{S\arabic{algocf}}
\renewcommand{\theHfigure}{S\arabic{figure}}
\renewcommand{\theHtable}{S\arabic{table}}
\renewcommand{\theHalgocf}{S\arabic{algocf}}
\section{Introduction}
\label{app:introduction}

This appendix provides implementation details of the rendering pipeline (\cref{app:implementation}) and the reconstruction pipeline (\cref{app:reconstruction}), including pseudocode, data layouts, and derivations.
It then reports additional results on performance and image quality (\cref{app:results}).
We follow the notation of the main paper, which \cref{tab:notation} summarizes.
We will release the source code of our implementation upon publication.

\begin{table}[t]
\centering
\small
\setlength{\tabcolsep}{4pt}
\begin{tabular}{@{}ll@{}}
\toprule
Symbol & Meaning \\
\midrule
$M$ & render tiles per macro-tile \\
$B_p$, $B_r$, $B_d$ & partitioning, rasterization, drain batch size \\
$P$, $W$ & threads per worker group, groups per unit \\
$K$ & mini-batches of a batch \\
$\tau$, $w$ & thread index, worker-group index \\
$\mathcal{S}$, $\mathcal{Q}$, $\mathcal{R}$ & local storage, tile queue, scheduling buffer \\
$\Sigma$, $L$, $l_k$ & covariance, Cholesky factor of $\Sigma^{-1}$, entries \\
$\boldsymbol{\mu}$, $\mathbf{p}$ & projected center of a Gaussian, pixel \\
$dx$, $dy$; $u_0$, $u_1$; $q$ & $\boldsymbol{\mu}-\mathbf{p}$; its projections on $L$; $u_0^2+u_1^2$ \\
$o_j$, $d_j$ & opacity of a Gaussian, $\partial\mathcal{L}/\partial\ln\alpha_j$ \\
$\varepsilon$ & transmittance threshold \\
$\mathbf{C}_i$, $T_i$ & color, transmittance of batch $i$ alone \\
$F_i$, $\mathbf{S}_i$ & transmittance before, color from batch $i$ on \\
$T_{\mathrm{end}}$, $\mathbf{b}$ & final transmittance, background color \\
$t_j$, $\alpha_j$ & local transmittance, opacity at a pixel \\
$\mathbf{c}_j$, $\widehat{w}_j$ & color, rounded blend weight \\
$\mathcal{L}$, $\mathbf{g}$, $\mathbf{g}_i$, $r$ & loss, its gradient, $F_i\mathbf{g}$, residual \\
$\theta$, $m$, $v$ & a parameter and its Adam moments \\
$\eta$, $\beta_1$, $\beta_2$, $\varepsilon_A$ & learning rate and Adam constants \\
$a$, $k$, $\rho$ & last updated step, lag, $\beta_1/\sqrt{\beta_2}$ \\
$\Phi$, $\Psi$ & sums of step factors over a lag \\
\bottomrule
\end{tabular}
\caption{Notation.}
\label{tab:notation}
\end{table}

\section{Rendering Pipeline}
\label{sec:impl}
\label{app:implementation}

\subsection{Packed SH coefficients}
\label{sec:fused-proj}

In the 32-byte SH representation of \cref*{M-sec:numerical} of the main paper, the constant term keeps 16~bits of luma and 15~bits per chroma channel of the YCoCg basis.
Each of the 15 higher-order coefficients is quantized to 6~bits of luma and 4~bits per chroma channel after square-root companding, with one scale per SH basis function and channel, the 99.99th percentile of its magnitude over Gaussians of opacity 0.005 or more.
Decoding inline from this representation shortens projection and SH evaluation by 35\% at 1080p and 30\% at 4K relative to streaming uncompressed FP16 coefficients.

\subsection{Segmented depth sort}
\label{sec:sort-impl}

Each stage of the sorting pipeline is one kernel, and the host launches the six back to back.

\begin{table}[h]
\centering
\small
\begin{tabular}{@{}llrl@{}}
\toprule
\multicolumn{2}{@{}l}{Stage} & Threads & Input \\
\midrule
Cascade & T0 & 64 & segments of 1 to 1{,}024 keys \\
        & T1 & 128 & segments of 1{,}025 to 2{,}048 keys \\
        & T2 & 256 & segments of 2{,}049 to 4{,}096 keys \\
        & T3 & 512 & segments of 4{,}097 to 8{,}192 keys \\
\multicolumn{2}{@{}l}{Run sort} & 512 & runs of 8{,}192 keys \\
\multicolumn{2}{@{}l}{Merge} & 256 & sorted runs of one segment \\
\bottomrule
\end{tabular}
\end{table}

\noindent\textbf{Cascade.}
Each tier is launched with one thread block per segment, and a block returns immediately unless its segment falls in the tier's size class.
A block sorts the 32-bit depth keys, the bit patterns of the positive FP32 depths, with their Gaussian ids in four 8-bit passes, using the block radix ranking of CUB~\cite{merrill2010revisiting}.
No tier reads what another writes, so the four overlap in execution through programmatic dependent launch~\cite{nvidia2025pdl}.

\noindent\textbf{Larger segments.}
The last tier does not sort a segment above 8{,}192 keys: it divides the segment into runs of 8{,}192 keys and lays out the binary tree of pairwise merges that joins them.
The run sort then sorts every run as a tier sorts a segment, and the merge executes the trees.
Each merge is partitioned by the GPU merge path~\cite{green2012gpu} into balanced subproblems, and a parent merge is scheduled on the device once its children are complete.
These two stages read what the stage before them wrote and run in sequence.
Each runs a fixed number of thread blocks that claim their work on the device, runs from a counter and merges from a queue, so the host never reads back how many large segments a frame holds.

\subsection{Batch rasterization}
\label{sec:raster-kernel}
\label{sec:gpu-mapping}

\begin{algorithm}[tp]
\caption{\textsc{ProcessingUnit}$(\mathit{batch},\mathit{pass})$: one processing unit, shared by the forward and the backward pass (\cref*{M-sec:fused-raster,M-sec:backward} of the main paper). Phase~1 assigns one Gaussian per thread, which hides memory latency, and worker group~0 builds the tile work queue $\mathcal{Q}$ before Phase~2.}
\label{fig:fused-raster}
\SetKwComment{Comment}{\quad$\triangleright$\ }{}
\KwIn{the $n\le B_r$ Gaussians of the batch, in depth order;\newline
      $K=\lceil n/P\rceil$ mini-batches of $P$ Gaussians;\newline
      $\mathit{pass}$ = forward or backward}
\BlankLine
\tcp{Phase 1: load and inline visibility}
\For(\Comment*[f]{groups stride over mini-batches}){$k \gets w$ \KwTo $K{-}1$ \textbf{step} $W$}{
    $j \gets kP + \tau$\Comment*[r]{one Gaussian per thread}
    \tcp{center, factors, color, in the macro-tile frame}
    \lIf{$j<n$}{$\mathcal{S}.\mathit{gauss}[j] \gets \textsc{Load}(\mathit{batch},j)$}
    \eIf{forward}{
        \tcp{$M$ bits: bit $t$ set if Gaussian $j$ overlaps tile $t$}
        $b \gets \textsc{OverlapTest}(\mathcal{S},j)$, or $0$ if $j\ge n$\;
        \tcp{thread $\tau$ receives the $P$ bits of tile $\tau$}
        $\mathcal{S}.\mathit{mask}[k][\tau] \gets \textsc{Transpose}(b)$\;
    }{
        \tcp{the mask of tile $\tau$ saved by the forward pass}
        $\mathcal{S}.\mathit{mask}[k][\tau] \gets \textsc{SavedMask}(k,\tau)$\;
    }
}
\textsc{Barrier}\;
\tcp{Tile work queue: thread $\tau$ of group 0 acts for tile $\tau$}
\If{$w = 0$}{
    \tcp{Gaussians of the batch at tile $\tau$}
    $h \gets \bigvee_{k=0}^{K-1} \mathcal{S}.\mathit{mask}[k][\tau]$\;
    $a \gets \textsc{Vote}(h \neq 0 \,\wedge\, \text{tile $\tau$ in the image})$\;
    \tcp{backward: keep tiles whose compositing reached the batch}
    \lIf{backward}{$a \gets a \wedge \textsc{Vote}(\text{tile $\tau$ composited})$}
    $\mathcal{Q} \gets \textsc{Compact}(a)$\;
}
\textsc{Barrier}\;
\tcp{Phase 2: per-tile work}
\While(\Comment*[f]{groups claim tiles dynamically}){$t \gets \textsc{Claim}(\mathcal{Q})$}{
    \textsc{RenderTile}$(t,\mathit{pass})$\Comment*[r]{\cref{fig:blend-detail}}
}
\end{algorithm}

\begin{algorithm}[tp]
\caption{\textsc{RenderTile}$(t,\mathit{pass})$: a worker group processes a render tile claimed in \cref{fig:fused-raster}, each thread at its pixel pair. Both passes walk the Gaussians of the tile in the same way, through the scheduling buffer $\mathcal{R}$ in drains of $B_d$, and differ in the routine applied to a drain. Saturation ends the walk early.}
\label{fig:blend-detail}
\SetKwComment{Comment}{\quad$\triangleright$\ }{}
\SetKwProg{Proc}{procedure}{}{end}
\KwIn{$\mathcal{S}$ and $K$ of \cref{fig:fused-raster};\newline
      backward: boundary state $F_i$, $\mathbf{S}_i$, $T_{\mathrm{end}}$ of the tile\newline
      and loss gradient $\mathbf{g}$, at the pixel pair}
\BlankLine
$T \gets 1$\Comment*[r]{transmittance within the batch}
\eIf{forward}{
    $\mathbf{C} \gets \mathbf{0}$\Comment*[r]{color of the batch}
}{
    $\mathbf{g}_i \gets F_i\,\mathbf{g}$;\quad $r \gets \mathbf{g}^\top(\mathbf{S}_i + T_{\mathrm{end}}\mathbf{b})$\Comment*[r]{residual}
}
$\mathcal{R} \gets \varnothing$\Comment*[r]{capacity $2B_d$}
\For(\Comment*[f]{mini-batches, in depth order}){$k \gets 0$ \KwTo $K{-}1$}{
    \tcp{Gaussians of mini-batch $k$ that overlap tile $t$}
    $\mathcal{R}.\textsc{Enqueue}(kP + \textsc{Compact}(\mathcal{S}.\mathit{mask}[k][t]))$\;
    \tcp{full drains, and the remainder after the last mini-batch}
    \While{$|\mathcal{R}| \ge B_d$ or ($k = K{-}1$ and $|\mathcal{R}| > 0$)}{
        $J \gets \mathcal{R}.\textsc{Dequeue}(\min(B_d, |\mathcal{R}|))$\;
        \eIf{forward}{
            \textsc{FrontToBackBlend}$(J)$\;
        }{
            \textsc{FrontToBackGradient}$(J)$\;
        }
        \tcp{every pixel of the tile is saturated}
        \lIf{$\textsc{Unanimous}(T < \varepsilon)$}{\textbf{break} out of both loops}
    }
}
\lIf{forward}{store $(\mathbf{C},T)$ as the partial $(\mathbf{C}_i,T_i)$ of tile $t$}
\BlankLine
\Proc{\textup{\textsc{FrontToBackBlend}}$(J)$}{
    \ForEach{Gaussian $j$ of $J$, in depth order}{
        $\alpha \gets \min(1,\,o_j\,2^{-q})$, set to $0$ below $1/255$\;
        $\widehat{w} \gets \alpha\,T$\;
        $\mathbf{C} \gets \mathbf{C} + \widehat{w}\,\mathbf{c}_j$;\quad $T \gets T - \widehat{w}$\;
    }
}
\Proc{\textup{\textsc{FrontToBackGradient}}$(J)$}{
    \ForEach{Gaussian $j$ of $J$, in depth order}{
        $\alpha \gets \min(1,\,o_j\,2^{-q})$, set to $0$ below $1/255$\;
        $\widehat{w} \gets \alpha\,T$\;
        $r \gets r - \widehat{w}\,\mathbf{g}_i^\top\mathbf{c}_j$\;
        \tcp{$d_j$ is zero if $\alpha=1$ or $T=0$}
        $d_j \gets \alpha\bigl(T\,\mathbf{g}_i^\top\mathbf{c}_j - r/(1-\alpha)\bigr)$;\quad $T \gets T - \widehat{w}$\;
        \tcp{sums over the pixels of the tile, \cref{eq:tile-sums}}
        \textsc{AtomicAdd}(gradients of $j$, sums)\;
    }
}
\end{algorithm}

\Cref{fig:fused-raster} gives the processing unit and \cref{fig:blend-detail} its walk over one render tile, and the backward branches are derived in \cref{app:backward}.
Every thread of a unit executes the pseudocode: thread $\tau\in[0,P)$ of worker group $w\in[0,W)$ holds its own copy of each variable, which in \cref{fig:blend-detail} is the value at the thread's pixel pair.
The following primitives act across threads:
\begin{itemize}
    \item \textsc{Barrier} waits until all worker groups of the unit arrive.
    \item $\textsc{Vote}(c)$ is the $P$-bit mask of the threads of a group at which $c$ holds, and $\textsc{Unanimous}(c)$ is true if $c$ holds at all of them.
    \item $\textsc{Compact}(\mathit{mask})$ returns the positions of the set bits of a mask.
    \item $\textsc{Transpose}(b)$ transposes the $P{\times}M$ bit matrix formed by the values $b$ of the threads of a group, so that thread $\tau$ receives column $\tau$.
    \item $\textsc{Claim}(\mathcal{Q})$ atomically removes one tile from the queue and returns it, or nothing once the queue is empty.
    \item \textsc{AtomicAdd} adds to a value that the threads of other groups and of other units update at the same time.
\end{itemize}
The loading mini-batch of $P=32$ Gaussians and the $M=32$ render tiles of a macro-tile are different axes that happen to have equal size in this configuration, which is what lets the transpose run within one worker group.
We use $W{=}10$ worker groups per unit for rendering and $W{=}8$ for training.

\noindent\textbf{Numerical format.}
The Cholesky factors $L=\bigl[\begin{smallmatrix} l_0 & 0\\ l_1 & l_2\end{smallmatrix}\bigr]$ are prescaled to absorb the change from natural to base-2 exponent and the tile-unit coordinate scale, $l'_k=l_k\,s\sqrt{\log_2(e)/2}$ for render tiles of $s$ pixels, so that the opacity at a pixel is the Gaussian's opacity times $2^{-q}$, evaluated with one FP16 base-2 exponential.
In the macro-tile frame of \cref*{M-sec:numerical} of the main paper, pixel coordinates satisfy $|x|\le4$ and $|y|\le2$.

\noindent\textbf{Blending and output.}
\textsc{FrontToBackBlend} is the blending recurrence of 3DGS, with opacity limited to one and the transmittance threshold $\varepsilon=10^{-4}$ tested after each drain.
Each unit writes its partials to a dense buffer indexed by batch and render tile, 16~KB per batch at four FP16 channels per pixel.

\section{Reconstruction Pipeline}
\label{app:reconstruction}

\subsection{Backward rasterization}
\label{app:backward}

\noindent\textbf{Boundary state.}
The training forward accumulates color in FP32 because FP16 accumulation swamps small increments and darkens the image, and because it would leave the backward pass a suffix that has already lost the terms it must subtract.
\Cref{alg:boundary} gives the compositing pass that completes the boundary state of a tile's batches.
The Gaussians of the batches behind the last one it composites receive zero gradient.

\begin{algorithm}[t]
\caption{\textsc{CompositeTraining}$(t)$: a worker group composites render tile $t$, each thread at its pixel pair, and completes the boundary state of the tile's batches.}
\label{alg:boundary}
\SetKwComment{Comment}{\quad$\triangleright$\ }{}
\KwIn{partials $(\mathbf{C}_i,T_i)$ of the tile's active batches\newline
      $i=0,\dots,n_b{-}1$, in depth order}
\KwOut{$F_i$, $\mathbf{S}_i$ of the composited batches $0,\dots,\mathit{last}$;\newline
       $T_{\mathrm{end}}$; the pixels of the image}
\BlankLine
$F\gets1$;\quad $\mathit{last}\gets n_b-1$\;
\For(\Comment*[f]{front to back}){$i\gets0$ \KwTo $n_b-1$}{
    $F_i\gets F$;\quad $F\gets F\,T_i$\;
    \tcp{every pixel of the tile is saturated}
    \lIf{$\textsc{Unanimous}(F<\varepsilon)$}{$\mathit{last}\gets i$; \textbf{break}}
}
$T_{\mathrm{end}}\gets F$;\quad $\mathbf{S}\gets\mathbf{0}$\;
\For(\Comment*[f]{back to front}){$i\gets\mathit{last}$ \KwTo $0$}{
    $\mathbf{S}\gets\mathbf{S}+F_i\,\mathbf{C}_i$;\quad $\mathbf{S}_i\gets\mathbf{S}$\;
}
image $\gets\mathbf{S}_0+T_{\mathrm{end}}\,\mathbf{b}$\;
\end{algorithm}

\noindent\textbf{Derivation.}
Consider one pixel and the Gaussians $j=1,\dots,n$ of batch $i$ in depth order, with local transmittance $t_1=1$, $t_{j+1}=t_j(1-\alpha_j)$ and blend weight $w_j=\alpha_jt_j$.
The pixel color is
\begin{equation}
 \mathbf{C}=\sum_{h<i}F_h\mathbf{C}_h+F_i\sum_{j}w_j\mathbf{c}_j+\sum_{h>i}F_h\mathbf{C}_h+T_{\mathrm{end}}\mathbf{b}.
\end{equation}
The first sum does not depend on batch $i$.
Every term behind Gaussian $j$, including the later batches and the background, carries the factor $(1-\alpha_j)$ through its transmittance, so
\begin{equation}
 \frac{\partial\mathbf{C}}{\partial\alpha_j}=F_it_j\mathbf{c}_j-\frac{1}{1-\alpha_j}\Bigl(F_i\!\sum_{j'>j}w_{j'}\mathbf{c}_{j'}+\sum_{h>i}F_h\mathbf{C}_h+T_{\mathrm{end}}\mathbf{b}\Bigr).
\end{equation}
Contracting with the loss gradient $\mathbf{g}$ and writing $\mathbf{g}_i=F_i\mathbf{g}$ gives \cref*{M-eq:backward} of the main paper, in which $\widehat{w}_j$ is the rounded value of $w_j$, with the residual
\begin{equation}
 r_j=\mathbf{g}^\top\Bigl(\mathbf{S}_i+T_{\mathrm{end}}\mathbf{b}\Bigr)-\sum_{j'\le j}w_{j'}\,\mathbf{g}_i^\top\mathbf{c}_{j'} .
\end{equation}
The recurrence of \cref*{M-eq:residual} of the main paper maintains it.
The derivative is that of the function the forward pass rendered, with rounding treated as the identity.
At $\alpha_j=1$ or $t_j=0$ the opacity-derived gradients are set to zero, and a Gaussian that the forward pass skipped at a pixel receives exactly zero there.

From $\alpha_j$ the chain continues to the inputs of the rasterizer.
Let $(dx,dy)=\boldsymbol{\mu}-\mathbf{p}$ be the offset of the center from the pixel and $l_k$ the prescaled factors of \cref{sec:raster-kernel}.
With $u_0=l_0\,dx+l_1\,dy$ and $u_1=l_2\,dy$, $q=u_0^2+u_1^2$, and $\alpha_j=o_j2^{-q}$ for opacity $o_j$, the gradient with respect to $\ln\alpha_j$ is $d_j=\alpha_j\,\partial\mathcal{L}/\partial\alpha_j$, and the per-Gaussian gradients are sums over the pixels $p$ of the tile:
\begin{align}
 \frac{\partial\mathcal{L}}{\partial o_j}&=\frac{1}{o_j}\sum_p d_j,\qquad
 \frac{\partial\mathcal{L}}{\partial l_0}\propto\sum_p u_0\,dx\,d_j,\nonumber\\
 \frac{\partial\mathcal{L}}{\partial l_1}&\propto\sum_p u_0\,dy\,d_j,\qquad
 \frac{\partial\mathcal{L}}{\partial l_2}\propto\sum_p u_1\,dy\,d_j,\label{eq:tile-sums}\\
 \frac{\partial\mathcal{L}}{\partial\mu_x}&\propto l_0\sum_p u_0\,d_j,\qquad
 \frac{\partial\mathcal{L}}{\partial\mu_y}\propto l_1\sum_p u_0\,d_j+l_2\sum_p u_1\,d_j,\nonumber
\end{align}
each with the constant $-2\ln2$, times the prescaling constant of \cref{sec:raster-kernel} for the factors and $1/s$ for the center, as the offsets are in units of render tiles.
A worker group adds these sums, and that of $\widehat{w}_j\,\mathbf{g}_i$ for the color, atomically to the gradients of the Gaussian's center, of its factors and opacity, and of its color, in five additions of two words each.

\subsection{Fused optimizer and stochastic rounding}
\label{app:adam}
\label{app:storage}

The fused pass applies Adam with $\beta_1=0.9$, $\beta_2=0.999$, and $\varepsilon_A=10^{-15}$.
A parameter is moved by the unrounded moments.
The gradient of the color with respect to the view direction contributes to the gradient of the mean, so the mean is updated last, after the SH coefficients.
Because the gradients do not persist, the densification statistics of 3DGS, the number of views and the norm of the screen-space gradient of the center, are accumulated in the same pass.

A store adds uniformly random bits to the bits it discards and then truncates, 16~bits for a BF16 moment and 13 for an FP16 parameter.
The bits are a hash of the run seed, the step, and the element index, so there is no generator state to store or to carry through densification.
The stored value is thus unbiased in the normal range of the format.
Widening an FP16 parameter to FP32 is exact, so a scene trained in this layout renders bit for bit like an FP32 scene holding the same values.

\begin{table*}[t]
\centering
\footnotesize
\setlength{\tabcolsep}{4pt}
\begin{tabular}{@{}lrrrrrrrrr@{}}
\toprule
Scene & HiGS & SHARP-GS & Faster-GS & FlashGS & TC-GS & Speedy-Splat & StopThePop & gsplat & 3DGS \\
\midrule
& \multicolumn{9}{c}{1080p throughput (FPS)}\\
\cmidrule(lr){2-10}
bicycle & 2053 & 806 & 640 & 582 & 468 & 415 & 374 & 357 & 163 \\
bonsai & 4219 & 2137 & 1481 & 1546 & 1443 & 1156 & 1036 & 979 & 557 \\
counter & 3546 & 1730 & 1370 & 1212 & 1242 & 925 & 789 & 725 & 343 \\
garden & 1901 & 839 & 583 & 591 & 466 & 418 & 375 & 361 & 222 \\
kitchen & 2632 & 1230 & 1021 & 852 & 916 & 679 & 615 & 554 & 295 \\
room & 3906 & 2008 & 1410 & 1466 & 1232 & 934 & 809 & 781 & 331 \\
stump & 2688 & 1235 & 996 & 912 & 682 & 625 & 560 & 535 & 371 \\
\midrule
& \multicolumn{9}{c}{4K throughput (FPS)}\\
\cmidrule(lr){2-10}
bicycle & 1309 & 756 & 465 & 508 & 356 & 296 & 218 & 205 & 60 \\
bonsai & 1934 & 1715 & 644 & 1121 & 839 & 608 & 481 & 450 & 167 \\
counter & 1698 & 1366 & 720 & 756 & 643 & 456 & 317 & 299 & 110 \\
garden & 1292 & 735 & 441 & 480 & 336 & 278 & 225 & 212 & 94 \\
kitchen & 1443 & 972 & 568 & 544 & 488 & 343 & 265 & 246 & 106 \\
room & 1736 & 1460 & 733 & 913 & 708 & 497 & 348 & 327 & 104 \\
stump & 1534 & 1087 & 615 & 796 & 492 & 407 & 341 & 312 & 140 \\
\bottomrule
\end{tabular}
\caption{Rendering throughput per scene on the pretrained 3DGS models of seven Mip-NeRF~360 scenes, in frames per second: $1000$ over the median frame time in milliseconds. TC-GS runs in its released integration with the Speedy-Splat rasterizer, StopThePop in its \texttt{vanilla} preset, which orders by depth per tile, and SHARP-GS with precomputed covariances and binning tiles of $32\times32$ pixels at 1080p and $64\times64$ at 4K. FlashGS fails on one kitchen camera at 4K, which is excluded from its median.}
\label{tab:rendering-full}
\end{table*}

\subsection{Deferred appearance updates}
\label{app:deferred}

\noindent\textbf{Schedule.}
At step $s$, a group with a visible Gaussian first brings the coefficients of all its Gaussians to step $s-1$ with \cref*{M-eq:catchup} of the main paper and then takes step $s$, with zero gradient for its culled members.
The forward and the backward pass of step $s$ both use the coefficients restored to step $s-1$.
Every lagging group is brought up to date before a densification, before a render that is not followed by an optimizer step, such as evaluation, and before a checkpoint, so that nothing outside the training step observes a lagging coefficient.
On \emph{garden}, a step's view culls on average 65\% of the Gaussians during densification and 66\% after it.

\noindent\textbf{Derivation of the closed form.}
In the notation of \cref*{M-eq:catchup} of the main paper, a zero-gradient Adam step $a+j$ scales $m$ by $\beta_1$ and $v$ by $\beta_2$ and moves a coefficient $\theta$ by
\begin{equation}
 -\frac{\eta}{1-\beta_1^{a+j}}\cdot\frac{\beta_1^{\,j}m}{\beta_2^{\,j/2}\sqrt{v}\big/\sqrt{1-\beta_2^{\,a+j}}+\varepsilon_A}.
 \label{eq:zero-step}
\end{equation}
Where $\varepsilon_A$ is negligible against $\sqrt{v}$, the $k$ steps sum to $-\eta\,\Phi\,m/\sqrt{v}$, and where it dominates, to $-\eta\,\Psi\,m/\varepsilon_A$.
\Cref*{M-eq:catchup} of the main paper reduces to each limit when the other term of its denominator is negligible, and to \cref{eq:zero-step} for $k=1$.
Between the limits it never underestimates the magnitude of the move, because the move of one step is concave and homogeneous in its two terms.
The overestimate is largest for a coefficient last updated at the first step of training, 4.5\%, and stays below 0.003\% from step 600 on.
With $\varepsilon_A=0$ the closed form coincides with the restoration of GS-Scale~\cite{lee2026gsscale}, which tabulates the factor $\eta\,\Phi$ for lags of up to 15~steps.
Where $\varepsilon_A$ dominates, that restoration differs from the sum of the steps by the ratio $\Phi/\Psi$, which is 0.31 to 0.32 at step 100 and within 0.5\% of one from step 5{,}000 on.
With $\bar\Phi(a)$ and $\bar\Psi(a)$ the same sums extended to infinity,
\begin{equation}
 \Phi=\bar\Phi(a)-\rho^{k}\,\bar\Phi(a+k),\qquad
 \Psi=\bar\Psi(a)-\beta_1^{k}\,\bar\Psi(a+k),
\end{equation}
so one table with two entries per step suffices.
The sums converge for $\beta_1^2<\beta_2$.
A restoration rounds the coefficients and moments once, instead of once per skipped step.
Over independently seeded roundings, the variance of a restored coefficient is 0.16 of that of a densely updated one at a lag of 8~steps and 0.05 at 38.

\subsection{Densification with spatial ordering}
\label{app:densification}

A densification step takes the Morton codes of the means, clamped to a grid of $2^{10}$ cells per axis fixed at the start of training over the bounding box of the initial means, widened by 10\%.
Its gather drops the pruned Gaussians and carries the moments with the parameters, and the step recorded for every group is then set to the current step.
From the same state and random stream, the resulting set of Gaussians and optimizer state equals that of gsplat's densification.

\section{Additional Results}
\label{app:results}

\subsection{Rendering performance}
\label{app:rendering-perf}

\begin{figure}[t]
\centering
\includegraphics[width=\linewidth]{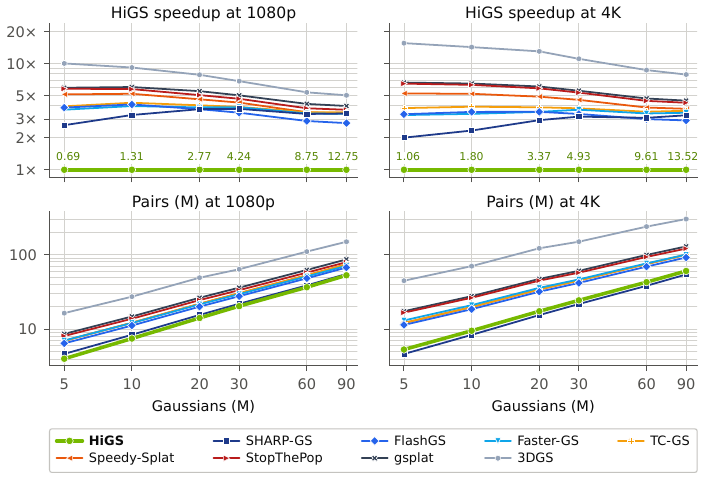}
\caption{\textbf{Scaling on the \emph{garden} scene}, reconstructed at six Gaussian budgets from 5M to 90M and rendered from the first camera of its test set. \emph{Top:} the speedup of HiGS over every rasterizer, the ratio of their median frame times over 400 frames at the same budget and resolution. The numbers above the line of HiGS are its frame times in milliseconds at each budget. \emph{Bottom:} the tile--Gaussian pairs that each method sorts: pairs with macro-tiles for HiGS and with $16\times16$ render tiles for the others ($16\times12$ for Faster-GS). The count of SHARP-GS differs in kind: it is taken over its binning tiles (\cref{tab:rendering-full}), from bounding rectangles and before the ellipse test that precedes its sort.}
\label{fig:garden-perf-scaling}
\end{figure}

\Cref{tab:rendering-full} expands the rendering comparison of \cref*{M-tab:fps-comparison} of the main paper per scene.
The margin of HiGS is largest on the scenes with the most Gaussians (\emph{bicycle}, \emph{garden}), where projection, binning, and sort take about half of the frame or more in the two baselines timed per stage, gsplat and FlashGS, and smallest against SHARP-GS at 4K on the small indoor scenes.

\begin{table}[t]
\centering
\footnotesize
\setlength{\tabcolsep}{2.5pt}
\begin{tabular}{@{}lrrrrrr@{}}
\toprule
& \multicolumn{3}{c}{1080p (M pairs)} & \multicolumn{3}{c}{4K (M pairs)}\\
\cmidrule(lr){2-4}\cmidrule(lr){5-7}
Scene & gsplat & Ellipse & HiGS & gsplat & Ellipse & HiGS\\
\midrule
bicycle & 5.16 & 3.97 & 2.15 & 11.08 & 7.01 & 2.95 \\
bonsai & 2.37 & 1.73 & 0.74 & 6.38 & 4.05 & 1.21 \\
counter & 3.71 & 2.62 & 1.05 & 10.25 & 6.21 & 1.80 \\
garden & 5.50 & 4.47 & 2.60 & 10.82 & 7.58 & 3.41 \\
kitchen & 4.79 & 3.74 & 1.81 & 11.32 & 7.72 & 2.69 \\
room & 3.03 & 2.06 & 0.77 & 8.83 & 5.19 & 1.39 \\
stump & 3.53 & 2.72 & 1.45 & 7.79 & 5.04 & 2.03 \\
\midrule
Mean & 4.01 & 3.04 & 1.51 & 9.49 & 6.11 & 2.21 \\
\bottomrule
\end{tabular}
\caption{Tile--Gaussian pair counts in millions on the seven Mip-NeRF~360 scenes, averaged over the test cameras. \emph{gsplat}: render-tile partitioning of the gsplat library~\cite{ye2024gsplat} at the evaluated revision, which bounds each Gaussian by an opacity-aware box~\cite{radl2024stopthepop}, with $16\times16$ render tiles. \emph{Ellipse}: the same boxes refined by per-row ellipse intersection~\cite{hanson2024speedysplat}, at the same tiles. \emph{HiGS}: the ellipse-refined test at macro-tile granularity, with $8\times8$ render tiles.}
\label{tab:pairs}
\end{table}

The pair reduction has two sources (\cref{tab:pairs}).
Refining gsplat's bounding boxes by ellipse intersection at its render tiles removes about a quarter of the pairs at 1080p and more than a third at 4K.
Moving the same test to macro-tile granularity removes half of what remains at 1080p and nearly two thirds at 4K, which makes it the larger source at both resolutions.
It is also the one that scales: from 1080p to 4K the macro-tile pairs grow by $1.3\times$ to $1.8\times$, on every scene less than gsplat's.

The segments quadruple at the same time, and the sorts respond differently.
A general-purpose segmented sort, CUB's over our 32-bit keys and macro-tile segments, slows by $1.9\times$ to $2.7\times$, on every scene more than the pairs grow.
Our sorting pipeline (\cref{sec:sort-impl}) slows by less than the pairs grow: its time changes by at most 2\% on five scenes and grows by up to $1.6\times$ on the two with the fewest pairs.
The global radix sort of gsplat slows by $2.9\times$ to $4.6\times$.

\begin{table}[t]
\centering
\footnotesize
\setlength{\tabcolsep}{2.5pt}
\begin{tabular}{@{}l rrrr rrrr@{}}
\toprule
& \multicolumn{4}{c}{1080p (ms)} & \multicolumn{4}{c}{4K (ms)} \\
\cmidrule(lr){2-5} \cmidrule(lr){6-9}
& \multicolumn{2}{c}{$8\times8$} & \multicolumn{2}{c}{$16\times16$} & \multicolumn{2}{c}{$8\times8$} & \multicolumn{2}{c}{$16\times16$} \\
\cmidrule(lr){2-3} \cmidrule(lr){4-5} \cmidrule(lr){6-7} \cmidrule(lr){8-9}
Scene & gsplat & Ours & gsplat & Ours & gsplat & Ours & gsplat & Ours \\
\midrule
bicycle & 3.51 & 0.49 & 2.81 & 0.62 & 8.13 & 0.76 & 4.88 & 0.87 \\
bonsai  & 1.41 & 0.24 & 1.02 & 0.29 & 4.94 & 0.52 & 2.22 & 0.50 \\
counter & 2.22 & 0.28 & 1.38 & 0.35 & 7.36 & 0.59 & 3.35 & 0.63 \\
garden  & 3.46 & 0.53 & 2.77 & 0.68 & 7.54 & 0.77 & 4.72 & 0.93 \\
kitchen & 2.66 & 0.38 & 1.81 & 0.50 & 7.90 & 0.69 & 4.07 & 0.79 \\
room    & 1.94 & 0.26 & 1.28 & 0.32 & 6.75 & 0.58 & 3.05 & 0.57 \\
stump   & 2.24 & 0.37 & 1.87 & 0.44 & 5.85 & 0.65 & 3.20 & 0.65 \\
\midrule
Mean & 2.49 & 0.36 & 1.85 & 0.46 & 6.92 & 0.65 & 3.64 & 0.71 \\
\bottomrule
\end{tabular}
\caption{Median frame time of gsplat and HiGS at two render-tile sizes on the seven Mip-NeRF~360 scenes.}
\label{tab:perf}
\end{table}

The decomposition is robust to the choice of render tile (\cref{tab:perf}).
Tiles of $16\times16$ pixels shorten gsplat's frame on every scene, by 17\% to 55\%.
HiGS has nothing comparable to gain, because its partition cost is already small: the larger tiles slow it on every scene at 1080p and change its frame time by $-4$\% to $+21$\% at 4K.
The comparison of the main paper therefore runs gsplat at $16\times16$, its faster setting.
At matched $8\times8$ tiles HiGS is $6.9\times$ faster at 1080p and $10.6\times$ at 4K.

\begin{table*}[t]
\centering
\footnotesize
\setlength{\tabcolsep}{3pt}
\begin{tabular}{@{}ll*{13}{r}@{}}
\toprule
 & & \multicolumn{9}{c}{Mip-NeRF 360} & \multicolumn{2}{c}{Tanks and Temples} & \multicolumn{2}{c}{Deep Blending} \\
\cmidrule(lr){3-11}\cmidrule(lr){12-13}\cmidrule(l){14-15}
 & Method & bicycle & flowers & garden & stump & treehill & bonsai & counter & kitchen & room & train & truck & drjohnson & playroom \\
\midrule
PSNR & 3DGS      & 25.62 & 21.93 & 27.81 & 26.88 & 22.97 & 32.48 & 29.21 & 31.56 & 31.80 & 21.23 & 25.54 & 29.48 & 30.18 \\
     & gsplat    & 25.63 & 21.95 & 27.82 & 26.93 & 22.97 & 32.41 & 29.20 & 31.66 & 31.81 & 21.29 & 25.53 & 29.49 & 30.16 \\
     & Faster-GS & 25.62 & 21.96 & 27.82 & 26.91 & 22.99 & 32.43 & 29.21 & 31.58 & 31.81 & 21.28 & 25.55 & 29.47 & 30.20 \\
     & HiGS      & 25.63 & 21.95 & 27.82 & 26.93 & 22.98 & 32.38 & 29.19 & 31.65 & 31.75 & 21.25 & 25.55 & 29.45 & 30.14 \\
\midrule
SSIM & 3DGS      & .7725 & .6179 & .8725 & .7798 & .6509 & .9482 & .9165 & .9336 & .9289 & .7952 & .8820 & .9060 & .9093 \\
     & gsplat    & .7751 & .6195 & .8728 & .7815 & .6522 & .9481 & .9166 & .9340 & .9291 & .7957 & .8820 & .9058 & .9092 \\
     & Faster-GS & .7742 & .6197 & .8728 & .7813 & .6518 & .9482 & .9167 & .9336 & .9290 & .7955 & .8823 & .9058 & .9094 \\
     & HiGS      & .7751 & .6196 & .8729 & .7817 & .6521 & .9482 & .9165 & .9337 & .9288 & .7954 & .8821 & .9055 & .9090 \\
\midrule
LPIPS & 3DGS      & .1652 & .3252 & .0727 & .1480 & .3155 & .1166 & .1427 & .0854 & .1525 & .1862 & .0956 & .1704 & .1595 \\
      & gsplat    & .1595 & .3227 & .0722 & .1452 & .3088 & .1164 & .1421 & .0849 & .1520 & .1851 & .0954 & .1699 & .1588 \\
      & Faster-GS & .1610 & .3228 & .0722 & .1455 & .3111 & .1162 & .1423 & .0851 & .1524 & .1856 & .0953 & .1699 & .1591 \\
      & HiGS      & .1595 & .3223 & .0721 & .1449 & .3074 & .1155 & .1418 & .0847 & .1513 & .1848 & .0948 & .1687 & .1576 \\
\midrule
Gaussians & 3DGS      & 4.60 & 2.70 & 3.58 & 4.12 & 3.04 & 1.06 & 1.03 & 1.57 & 1.26 & 1.04 & 2.01 & 3.14 & 1.87 \\
(M)       & gsplat    & 4.78 & 2.76 & 3.64 & 4.15 & 3.11 & 1.09 & 1.06 & 1.60 & 1.31 & 1.06 & 2.04 & 3.27 & 1.94 \\
          & Faster-GS & 4.73 & 2.75 & 3.62 & 4.12 & 3.08 & 1.08 & 1.05 & 1.59 & 1.30 & 1.05 & 2.04 & 3.25 & 1.92 \\
          & HiGS      & 4.79 & 2.75 & 3.64 & 4.13 & 3.12 & 1.09 & 1.06 & 1.60 & 1.30 & 1.06 & 2.04 & 3.27 & 1.93 \\
\midrule
Memory & 3DGS      & 6.57 & 3.90 & 5.13 & 5.90 & 4.32 & 1.65 & 1.79 & 2.55 & 2.06 & 1.57 & 2.90 & 4.72 & 2.74 \\
(GiB)  & gsplat    & 6.98 & 4.06 & 5.34 & 6.06 & 4.57 & 1.69 & 1.66 & 2.43 & 2.00 & 1.60 & 3.01 & 4.83 & 2.89 \\
       & Faster-GS & 10.34 & 6.05 & 7.96 & 9.14 & 6.76 & 2.48 & 2.39 & 3.53 & 2.93 & 2.33 & 4.49 & 7.20 & 4.28 \\
       & HiGS      & 3.52 & 2.07 & 2.72 & 3.06 & 2.32 & 1.04 & 1.04 & 1.29 & 1.13 & 0.84 & 1.55 & 2.53 & 1.50 \\
\midrule
Time & 3DGS      & 759.9 & 537.9 & 659.9 & 600.4 & 571.6 & 551.5 & 663.2 & 792.1 & 721.7 & 373.7 & 425.8 & 858.3 & 672.1 \\
(s)  & gsplat    & 481.0 & 296.3 & 404.8 & 395.4 & 324.6 & 182.0 & 205.7 & 274.9 & 219.9 & 177.1 & 250.6 & 348.5 & 239.2 \\
     & Faster-GS & 191.1 & 122.7 & 170.3 & 160.1 & 130.6 & 84.1 & 93.8 & 126.9 & 95.1 & 69.6 & 101.1 & 139.4 & 97.9 \\
     & HiGS      & 60.1 & 44.4 & 61.2 & 47.5 & 46.3 & 38.4 & 44.9 & 58.9 & 41.6 & 33.3 & 39.7 & 44.6 & 35.6 \\
\bottomrule
\end{tabular}
\caption{\textbf{Reconstruction per scene}, mean over ten runs and, for time and memory, over three. Measures as in \cref*{M-tab:training} of the main paper, with the final number of Gaussians in millions.}
\label{tab:training-scenes}
\end{table*}

On the \emph{garden} scene of Mip-NeRF~360 (\cref{fig:garden-perf-scaling}), HiGS leads every method at every budget and at both resolutions.
From the smallest model to the largest, 18 times the Gaussians, its frame time grows 18.4$\times$ at 1080p and 12.8$\times$ at 4K.
Its lead over gsplat narrows as the model grows, because per-Gaussian projection, which every method pays, takes a growing share of the frame: at 1080p, 29\% of the HiGS frame at 5M and 38\% at 90M.
Among the per-tile pipelines, frame time orders the methods as the pair count does: in both rows of \cref{fig:garden-perf-scaling}, 3DGS stands above gsplat and StopThePop at every budget, and these above FlashGS, Faster-GS, TC-GS, and Speedy-Splat, which form the fewest pairs.
SHARP-GS is the fastest baseline for the two smallest models, then falls behind FlashGS, and the lead of HiGS over it widens with the model.
Its binning and rasterization are the cause: from 5M to 90M they take 21.7$\times$ longer at 4K, against 10.2$\times$ for those of HiGS, while projection grows alike in both.

\begin{table}[t]
\centering
\footnotesize
\setlength{\tabcolsep}{2.5pt}
\begin{tabular}{@{}l rrr rrr@{}}
\toprule
 & \multicolumn{3}{c}{against gsplat} & \multicolumn{3}{c}{against the test set} \\
\cmidrule(lr){2-4} \cmidrule(lr){5-7}
Method & PSNR & SSIM & LPIPS & PSNR & SSIM & LPIPS \\
\midrule
Ours              & 67.16 & .9999 & .0000 & 28.94 & .8721 & .1285 \\
Ours, packed SH   & 55.60 & .9995 & .0007 & 28.93 & .8718 & .1286 \\
FlashGS           & 49.74 & .9983 & .0010 & 28.92 & .8713 & .1282 \\
Faster-GS         & 73.86 & 1.0000 & .0000 & 28.94 & .8721 & .1287 \\
Speedy-Splat      & 94.43 & 1.0000 & .0000 & 28.94 & .8721 & .1287 \\
StopThePop        & 94.37 & 1.0000 & .0000 & 28.94 & .8721 & .1287 \\
3DGS              & 75.04 & 1.0000 & .0000 & 28.94 & .8721 & .1287 \\
gsplat            & ---   & ---   & ---   & 28.94 & .8721 & .1287 \\
\bottomrule
\end{tabular}
\caption{Image quality of rendering the pretrained 3DGS models of seven Mip-NeRF~360 scenes, mean over scenes. \emph{Left:} against the FP32 gsplat renderer at 1080p, which isolates the renderer with the model held fixed. \emph{Right:} against the test set at the standard resolution of each scene.}
\label{tab:iq-comparison}
\end{table}

\begin{figure}[t]
\centering
\includegraphics[width=\linewidth]{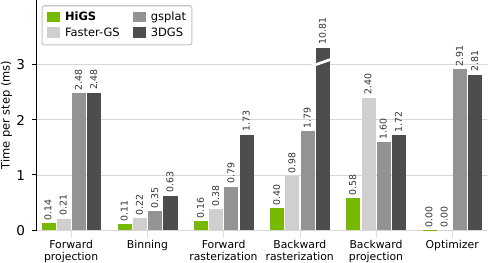}
\caption{\textbf{Stages of a steady-state training step}: mean time of each stage (in ms) over the steps after densification ends and over the 13 scenes of \cref*{M-tab:training} of the main paper, one run per scene, timed with CUDA events at the stage boundaries. \emph{Forward projection} includes SH evaluation, \emph{binning} the depth sort, \emph{forward rasterization} the compositing pass of HiGS, and \emph{backward projection} the SH gradient and, for HiGS and Faster-GS, the fused optimizer, so only gsplat and 3DGS have an optimizer stage. Not shown are the loss with its gradient, computed by the same code for every method, and the remainder of a step outside these stages, on average 0.25~ms and 0.01~ms.}
\label{fig:training-stages}
\end{figure}

\subsection{Training performance}
\label{app:training-perf}

HiGS is the fastest in every stage of a steady-state training step on every scene (\cref{fig:training-stages}).
Its step takes 1.64~ms on average, against 4.45~ms for Faster-GS, 10.20~ms for gsplat, and 20.44~ms for 3DGS.
Binning and the two rasterization passes, whose work is the tile--Gaussian pairs, are 2.3$\times$ to 19.5$\times$ faster than in the other methods, and in 3DGS backward rasterization alone is more than half of the step.
The two projections and the optimizer work per Gaussian, and their time follows how a method treats its parameters.
Faster-GS, whose fused optimizer updates every Gaussian at every step, spends 3.7$\times$ the time of HiGS on them.
Both gsplat and 3DGS spend 9.8$\times$, as they also concatenate their SH coefficients, write a full parameter gradient, and run Adam as a separate pass.
Backward projection with the fused optimizer is 35\% of a HiGS step and 54\% of a Faster-GS step.

In a steady-state training step on \emph{garden}, the fused optimizer runs at 80\% of the peak memory bandwidth of the GPU, whereas forward and backward rasterization are bound by instruction throughput rather than memory.
Each step restores the appearance of 13.1\% of the Gaussians, those in groups that were not updated at the previous step, and 96.0\% of them lag by at most 16~steps.

Per scene (\cref{tab:training-scenes}), HiGS uses 1.6$\times$ to 1.9$\times$ less memory than the most memory-efficient of the other methods and ends with the number of Gaussians of gsplat to within 0.02M.
Its PSNR is within 0.10~dB of 3DGS and gsplat on every scene, above on some scenes and below on others.

\begin{figure}[t]
\centering
\includegraphics[width=\linewidth]{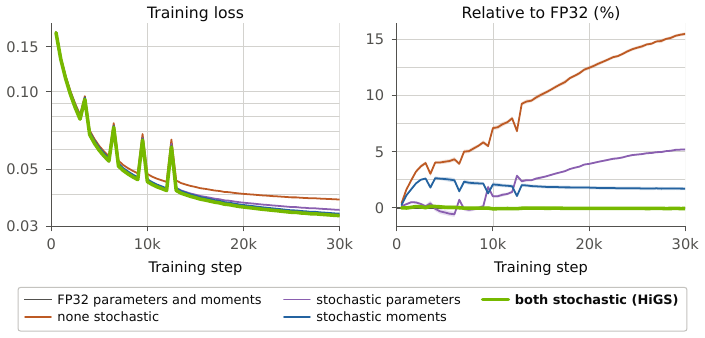}
\caption{\textbf{Training loss with and without stochastic rounding} on the \emph{garden} scene, for FP16 parameters and BF16 moments that are each rounded stochastically or to nearest. \emph{Left:} the loss over windows of 500~steps, mean over ten runs. \emph{Right:} the same loss relative to FP32 parameters and moments, each run against the FP32 run of the same seed, with the 95\% interval of the mean.}
\label{fig:rounding-loss}
\end{figure}

The training loss (\cref{fig:rounding-loss}) follows through the run what the image metrics of \cref*{M-tab:ablation} of the main paper measure at its end.
With every store rounded to nearest, the loss over the last 5{,}000 steps is 14.9\% above that of FP32 parameters and moments, the second row of that table, and the gap is still widening.
The drift belongs to the moments: with only the moments rounded to nearest the gap still widens, to 5.0\% over the last 5{,}000 steps.
A second moment decays by the factor $\beta_2$ at each step, a change of 0.1\% against a spacing of BF16 values of 0.4\% to 0.8\%, so the second moment of a parameter that is updated at every step is rounded back to its previous value and can only grow.
It then records the largest squared gradient the parameter has met rather than a recent average, and the steps that Adam divides by it shrink as training proceeds.
With only the parameters rounded to nearest the gap settles instead at a constant 1.7\%, the cost of the updates below half the spacing of FP16 values that such a store discards.
The two do not simply add: rounding both to nearest costs about twice the sum of the two.
With both rounded stochastically the loss stays within 0.2\% of FP32 over the whole run, a remainder we attribute to the noise that every store adds.
The excursions of the relative loss before step 15{,}000 follow the opacity resets.

\begin{table}[t]
\centering
\footnotesize
\setlength{\tabcolsep}{4pt}
\begin{tabular}{@{}ll ccc@{}}
\toprule
          &               & Mip-NeRF & Tanks and & Deep \\
          & Method        & 360      & Temples   & Blending \\
\midrule
Time      & SAD-GS        & 31.2  & 19.0  & 15.6 \\
(s)       & HiGS + SAD-GS & \phantom{0}7.1 & \phantom{0}4.6 & \phantom{0}4.1 \\
\midrule
PSNR      & SAD-GS        & 27.15 & 22.50 & 28.97 \\
(dB)      & HiGS + SAD-GS & 27.29 & 22.52 & 29.08 \\
\midrule
Gaussians & SAD-GS        & 3.06  & 1.35  & 0.88 \\
(M)       & HiGS + SAD-GS & 2.90  & 1.29  & 0.78 \\
\bottomrule
\end{tabular}
\caption{\textbf{Reconstruction with the densification of SAD-GS}: the harness of the main paper with the densification and training settings of SAD-GS in place of the 3DGS policy, for 3{,}000 steps. \emph{HiGS + SAD-GS} rasterizes with antialiasing. Time excludes the structure analysis of the training images that precedes training in both methods. Mean over the scenes of each dataset and three runs.}
\label{tab:structure-aware}
\end{table}

HiGS is not tied to the densification policy of 3DGS and combines with more recent ones.
Trained for 3{,}000 steps with the structure-aware densification of SAD-GS~\cite{lyu2026sadgs}, it is 3.8$\times$ to 4.4$\times$ faster than SAD-GS itself at an equivalent PSNR (\cref{tab:structure-aware}).
SAD-GS lowers the number of steps of a run and HiGS the cost of each.

\subsection{Image quality}
\label{sec:quality}

Against the test set, the differences between the rasterizers lie below the residual error of the trained model (\cref{tab:iq-comparison}).
Against the FP32 gsplat renderer they span tens of decibels, and the lower levels trace to specific design choices.
Speedy-Splat and StopThePop reproduce the reference to numerical precision.
The fixed $3\sigma$ tile bound of 3DGS can omit contributions above the $1/255$ threshold, as StopThePop notes, Faster-GS tests termination after compositing and does not clamp opacity from above, and FlashGS writes an 8-bit pixel buffer.
At 8~bits per channel, the level of HiGS corresponds to a root-mean-square error of about 0.1 code values.

\end{document}